%% file: main.tex
\documentclass[sigconf, nonacm]{acmart}

\usepackage{bm}
\usepackage{subfigure}
\usepackage{graphicx}
\usepackage{multirow}
\usepackage{enumitem}

\newcommand\vldbdoi{XX.XX/XXX.XX}

\begin{document}
\title{Exploring the Robustness of Decentralized Training for Large Language Models}

\author{Lin Lu$^*$}
\affiliation{%
  \institution{Hubei Engineering Research Center on Big Data Security, School of Cyber Science and Engineering, Huazhong University of Science of Technology}
  \city{Wuhan}
  \state{China}
}
\email{loserlulin@hust.edu.cn}

\author{Chenxi Dai$^*$}
\affiliation{%
  \institution{Hubei Engineering Research Center on Big Data Security, School of Cyber Science and Engineering, Huazhong University of Science of Technology}
  \city{Wuhan}
  \state{China}
}
\email{dcx001@hust.edu.cn}

\author{Wangcheng Tao}
\affiliation{%
  \institution{School of Cyber Science and Engineering, Huazhong University of Science of Technology}
  \city{Wuhan}
  \state{China}
}
\email{dangmai@hust.edu.cn}

\author{Binhang Yuan}
\affiliation{%
  \institution{Hong Kong University of Science of Technology}
  \city{Hong Kong}
  \state{China}
}
\email{biyuan@ust.hk}

\author{Yanan Sun}
\affiliation{%
  \institution{College of Computer Science, Sichuan University}
  \city{Sichuan}
  \state{China}
}
\email{ysun@scu.edu.cn}

\author{Pan Zhou}
\affiliation{%
  \institution{Hubei Engineering Research Center on Big Data Security, School of Cyber Science and Engineering, Huazhong University of Science of Technology}
  \city{Wuhan}
  \state{China}
}
\email{panzhou@hust.edu.cn}

\begin{abstract}
Decentralized training of large language models has emerged as an effective way to democratize this technology. However, the potential threats associated with this approach have not been carefully discussed, which would hinder the development of decentralized training infrastructures. This paper aims to initiate discussion towards this end by exploring the robustness of decentralized training from three main perspectives. First, we demonstrate the vulnerabilities inherent in decentralized training frameworks in terms of hardware, data, and models. Second, we highlight the fundamental difference between decentralized foundation model training and vanilla federated learning, where the security techniques employed in federated learning cannot be applied directly. Third, we discuss the essential components required for a robust and efficient decentralized training framework and present a case study by modeling a concrete threat model. Our objective in this vision paper is to emphasize the importance of addressing security concerns in the context of decentralized training for large language models.
\end{abstract}

\maketitle



\section{Introduction}
\label{Introduction}


Large language models (LLMs)~\cite{brown2020language, touvron2023llama, zhang2022opt, workshop2022bloom} have shown exceptional accuracy in numerous natural language processing tasks, thus gaining widespread acceptance and usage~\cite{cui2023chatlaw, singhal2023large, shen2023hugginggpt}. However, to improve accuracy in various domains, LLMs have expanded aggressively in terms of model scale and pre-train data volumes, resulting in time- and cost-intensive training processes \citep{bommasani2021opportunities, kaddour2023challenges, zhao2023survey}. For example, the state-of-the-art Falcon-180B~\cite{falcon180b} model has 180 billion parameters trained on 3.5 trillion tokens. Given the intensive computational load, sophisticated parallel strategies must be leveraged to speed up and scale out the training procedure~\cite{huang2019gpipe,li2020pytorch,narayanan2019pipedream,narayanan2021efficient,rajbhandari2020zero}.



A promising direction to democratize the training of large language models is through decentralized training \citep{diskin2021distributed, ryabinin2023swarm,yuan2022decentralized}, which presents a substantial solution to alleviate this resource-intensive challenge. 
On the other hand, these decentralized training frameworks are primarily based on \textit{model parallelism} (e.g, \textit{pipeline parallelism}~\cite{huang2019gpipe,narayanan2019pipedream}), supplemented by data parallelism. 
These parallel paradigms require communication of \textit{activations} during forward propagation and \textit{corresponding gradients} during backward propagation, which is fundamentally different from vanilla federated learning (FL) that only requires synchronization of \textit{model gradients} in a data parallel paradigm.
As a result, the potential risks and vulnerabilities associated with such decentralized training have not been formally discussed, to the best of our knowledge. 




The most relevant technique discussed in the data management and machine learning communities is secure aggregation in FL, which limits its scope under the data parallel communication paradigm ~\cite{fang2022bridge, karimireddy2021learning, farhadkhani2022byzantine, tao2023byzantineresilient}. In such scenarios, when malicious gradient values arise, the parameter server employs resilient gradient aggregation methods. These methods mainly employ outlier detection algorithms, such as the voting mechanism and the bucketing mechanism, to mitigate the impact of these malicious gradient values on the global model. On the other hand, safety issues under the scope of model parallelism are mostly unexplored, where the communication of activations and the corresponding gradients demands different approaches for malicious detection and defense.

\label{ThreeQuestions}
Therefore, in this paper, we initiate the discussion of three fundamental questions about the robustness of decentralized training, particularly in the context of pipeline parallelism. For each question, we give our answer and make a detailed explanation: 

\begin{itemize}[topsep=5pt, leftmargin=*]
    \item \textit{\textbf{Q1: What types of threat may occur in decentralized training? How will they influence the statistical efficiency of the training?}}
    To assess the vulnerability and sensitivity of decentralized training, we have analyzed three potential threats: hardware failures, privacy inference attacks, and poisoning attacks. These threats represent three kinds of malicious attackers with escalating attack capabilities. We have investigated the feasibility of these attack forms under the scenario of decentralized training and emphasized their potential consequences.

    \item \textit{\textbf{Q2: Can the existing defense methods in traditional distributed learning (i.e., FL) be applied to decentralized training based on pipeline parallelism directly?}} The short answer is No --- We enumerate two primary distinctions between decentralized training and FL, which underscore the structural differences between pipeline parallelism and data parallelism techniques. These distinctions lead to notable discrepancies in the threats encountered with the decentralized training framework compared to previous security challenges of distributed systems. Consequently, defense algorithms tailored for traditional data parallelism techniques cannot be directly employed.

    \item \textit{\textbf{Q3: From what perspectives can we enhance the robustness of the decentralized training framework?}} Based on the potential threats mentioned above, our study focuses on determining the fundamental components necessary for a robust and efficient decentralized training framework. We also explain the organizational architecture of these components to effectively mitigate the aforementioned threats.
\end{itemize}

We also present a case study illustrating a straightforward and potent poisoning attack method targeting the forward and backward data propagation processes within a decentralized training framework. This case study emphasizes the urgency of addressing this issue as such poisoning attacks can profoundly impede model convergence and compromise model performance. To encounter this attack, we propose a relatively robust and efficient training framework. Additionally, we validate the effectiveness of both our attack and defense strategies through experimental verification.

The primary objective of this paper is to examine the potential threats inherent in decentralized training frameworks and propose possible defense methods to tackle these challenges. We hope that the discussion presented in this study will garner substantial attention from researchers specializing in related fields.

\vspace{-0.5em}
\section{Background}

\begin{figure}[t]
    \centering
    \includegraphics[width=\linewidth,page=1]{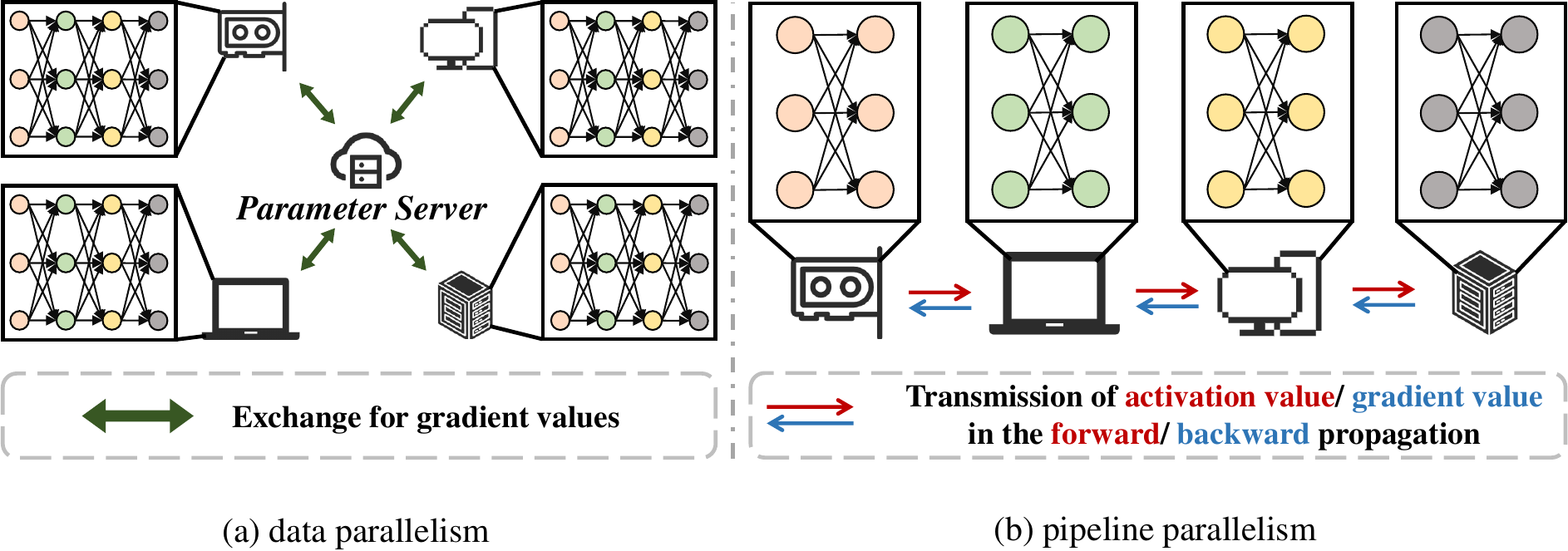}
    \caption{A comparison of model layer segmentation in data parallelism and pipeline parallelism.}
    \label{figure-1}
    \vspace{-1.5em}
\end{figure}

\noindent\textbf{Parallel training for LLMs.} To distribute the training computation of large language models over thousands of compute devices (usually GPUs), different categories of parallel strategies have been proposed.
\textit{Data parallelism} partitions the mini-batch by training samples to distribute the computation load, where each GPU holds a local model replica for forward and backward propagations and communicates the gradients for synchronization, usually by a parameter server or an \texttt{AllReduce} operation~\cite{li2020pytorch}.
FL~\cite{mcmahan2017communication, konevcny2016federated, bonawitz2019towards} is mainly based on data parallelism. 
Figure \ref{figure-1}(a) illustrates an example of data parallelism with 4 workers. They send gradients computed from their local datasets to the parameter server and receive aggregated gradients to update their local models \citep{zhang2015deep, reddi2016aide}.
\textit{Pipeline parallelism} partitions the training computation into multiple stages as a pipeline, where each GPU handles one stage.
Figure \ref{figure-1}(b) provides an illustration of pipeline parallelism, in which the model is partitioned into distinct sub-models, and each computational device handles a specific subset of model layers \citep{huang2019gpipe, narayanan2019pipedream, yang2021pipemare}. In contrast to data parallelism, pipeline parallelism requires fewer communication exchanges and optimizes the utilization of computational resources \citep{shoeybi2019magatronlm, rasley2020deepspeed, baines2021fairscale}. Due to these advantages, pipeline parallelism has become the main technique for decentralized training.



\noindent\textbf{Decentralized training.} Decentralized training strategies have emerged as practical means to facilitate collaborative training of LLMs among multiple contributors, thereby enhancing the democratization of the training process. \citep{yuan2022decentralized} initially investigates the decentralized training for large foundation models using model parallelism. Subsequently, \citep{krause2023swarm} and \citep{wang2023cocktailsgd} accomplish billion-scale training on heterogeneous devices with slow interconnect. Similarly, \citep{tang2023fusionai} aims to leverage vast untapped consumer-level GPUs.

\noindent\textbf{Robustness of the decentralized training.} While the security problems in decentralized training have been mentioned in previous works \citep{tang2023fusionai, borzunov2022training}, no systematic research has studied this issue extensively. Existing research focuses mainly on ensuring seamless pipeline operations \citep{athlur2022varuna, thorpe2023bamboo, jang2023oobleck}. However, these discussions face a prevalent limitation. They only discuss machine failures, neglecting the vulnerability of decentralized training to various imperceptible security risks.

\section{Potential Threats}
\label{Sec-PotentialThreats}
This section aims to address \textbf{Q1} by discussing the potential threats in decentralized training, including hardware failures, privacy inference attacks, and poisoning attacks. We observe that these three threat forms represent attackers with increasing capabilities. The weakest attacker can cause hardware failures without access to the training datasets or the transmitting values. A stronger attacker can steal the data during training, while the strongest attacker can manipulate the transmitting activation values or gradient values.

\subsection{Hardware Failures}
Hardware failures are common in distributed learning systems. For instance, Meta AI experienced numerous hardware failures while training OPT \citep{zhang2022opt}, resulting in over one hundred restarts in their compute cluster. In the event of a hardware failure, decentralized training keeps all training resources inactive until repairs are made, leading to significant resource wastage. 

This matter has garnered significant attention as a problem of fault tolerance and frequent interruptions. Relevant research focuses on maintaining model throughput while enabling automatic recovery from hardware failures. For instance, Varuna \citep{athlur2022varuna} introduces job morphing, allowing for the reconfiguration of the training job. Similarly, Bamboo \citep{thorpe2023bamboo} and Oobleck \citep{jang2023oobleck} facilitate the use of backup computing resources in case of hardware failures, ensuring a seamless training process.

However, several challenges remain unsolved. For example, the aforementioned approaches experience additional overhead when restarting the training process or require additional computing resources, which leads to the inefficient utilization of computing resources and contradicts the original objective of decentralized training strategies. A more serious problem is how to ensure the normal operation of the pipeline in the face of large-scale and high-frequency hardware failures.

\subsection{Privacy Inference Attacks}
LLMs have found widespread application in various domains, such as healthcare \citep{xiong2023doctorglm, singhal2022large} and law \citep{cui2023chatlaw, huang2023lawyer}. Data privacy concerns in these domains make the fine-tuning process of LLMs vulnerable to privacy inference attacks. Decentralized training frameworks are particularly susceptible to privacy inference attacks due to the frequent exchange of data and the inherent openness of distributed environments. For instance, gradient values enable an adversary to obtain training inputs with only a few iterations, as highlighted in \citep{zhu2019deep, aono2017privacy}. \citep{zhao2020idlg} introduces an approach that achieves 100\% accuracy in extracting ground-truth labels from the gradients.

Furthermore, in addition to directly accessing the original data, there are studies \citep{ateniese2015hacking, hitaj2017deep} that focus on properties unrelated to the characteristic features of the class. These studies show that an attacker, armed with auxiliary training data labeled with the desired property, can deduce valuable information that was previously unknown. Whether through direct or indirect means, privacy inference attacks pose a risk in decentralized training frameworks, potentially exposing sensitive content in the training datasets.

\subsection{Poisoning Attacks}
\label{Subsec-PoisoningAttacks}
In contrast to the act of stealing information in privacy inference attacks, poisoning attacks enable attackers to manipulate data transmission between stages. Depending on the attacker's objectives, poisoning attacks can be categorized as targeted attacks or untargeted attacks. Targeted attacks hinder the model's convergence by freely manipulating transmitting values, whereas untargeted attacks aim to inject backdoors into the global model.

Previous studies \citep{cao2019understanding, tolpegin2020data} thoroughly investigate the detrimental impact of untargeted attacks on the convergence of the global model in distributed systems. However, these studies were either limited to FL scenarios or only involved poisoning datasets by tampering with the corresponding labels. We evaluate the vulnerability of decentralized training to untargeted attacks in Section \ref{Sec-Differences}, providing an explanation for why decentralized training frameworks are more susceptible to such attacks compared to FL.

In the case of targeted attacks, a significant distinction arises from the inherent assumption of absolute security regarding the data providers in decentralized training. Nevertheless, several studies \citep{li2021backdoor, hong2022handcrafted} demonstrate the feasibility of implanting backdoors without access to the original data. Since the decentralized training framework involves frequent transfer and update of gradients, these attacks can be applied to decentralized learning as well. The frequent data exchange of decentralized training provides a new form of poisoning attacks, that is tampering with the activation values or gradient values. We show the possible consequences of this new untargeted poisoning attack form in our case study.

\section{Limitation of Secure Aggregate in FL}
\label{Sec-Differences}

In this section, we aim to address \textbf{Q2}. We posit that the direct application of current security methods in FL to decentralized training encounters significant challenges for the following reasons.

\subsection{Inherent Serial Characteristic}
Decentralized training frameworks primarily rely on pipeline parallelism as the main training technique. However, due to limited computational resources, a majority of training initiators only deploy one pipeline. This constraint results in an inherent serial characteristic within decentralized training frameworks, impeding the direct application of existing methods in two critical aspects.

\noindent\textbf{Lack of comparable values.}
In traditional FL, each worker possesses a complete copy of the global model. Privacy-preserving techniques, such as secure multiparty computation \citep{bonawitz2017practical} or secret-sharing-based methods \citep{bonawitz2017practical}, are used to prevent privacy inference attacks. To mitigate poisoning attacks, outlier detection algorithms, like the voting mechanism \citep{melnyk2018byzantine, datar2022byzantine, wang2019byzantine} and bucketing mechanism \citep{karimireddy2021byzantinerobust, zhu2023byzantinerobust, allouah2023fixing} can be employed to filter the Byzantine workers. 

However, during decentralized training, each stage in the pipeline can solely receive activation values or gradient values from the preceding stage. Due to the lack of comparable values, directly applying outlier detection algorithms or other privacy-preserving methods is not feasible. Although some training initiators try to solve this problem by adding more pipelines \citep{li2021chimera, narayanan2021efficient, jang2023oobleck}, striking a balance between computing resource utilization and obtaining an adequate number of comparable values is challenging.

\noindent\textbf{Heavy dependence on the predecessor stage.}
Each stage in decentralized training relies exclusively on the preceding stage due to the absence of a central server. In the context of poisoning attacks, if a stage becomes malicious, the remaining stages will remain unaware and mistakenly treat the malicious stage as honest. Furthermore, once a malicious stage manipulates the transmitting values, the subsequent stage cannot detect this malicious behavior and can only propagate the tampered data.

To illustrate, we consider the scenario where a malicious stage transmits an all-zero vector to the next stage. The honest stage is unable to determine if the value has been maliciously tampered with by the available algorithm and must rely on the preceding stage. In Subsection \ref{Subsec-PoisoningAttacks}, we extensively discuss the dangers associated with poisoning attacks. However, in real training scenarios, such malicious alterations to the transmitting values will be considerably less apparent, but the resulting harm can still be substantial.

\subsection{Change of Exchange Object and Frequency}


Compared to data exchange between the parameter server and workers, the exchange objects and frequency have changed a lot in decentralized training. In terms of the exchange object, stages should additionally transmit activation values in the forward propagation. Compared to gradient values, activation values vary more with the training data. As a result, the average-value-based resilient aggregation method cannot ensure the accuracy of training.

On the other hand, the parameter server only exchanges with the workers once during each iteration. However, the number of data exchanges in the decentralized training relies on the number of stages. The unknown target of the attacker requires a robust algorithm in every data exchange, which undoubtedly extends the training time and greatly reduces the training efficiency.

\section{Robust Decentralized Training}

In this section, our attention is centered on \textbf{Q3}. Here, we delineate vital components necessary for a robust and efficient decentralized training framework. In addition, we analyze the associated challenges. From a theoretical perspective, we discuss the viability of existing defense algorithms in mitigating privacy inference attacks and poisoning attacks. Furthermore, we underscore the imperative of fast recovery from a systematic viewpoint.

\subsection{Privacy Preservation}
Privacy-preserving methods, particularly in FL, have gained significant traction in diverse areas of machine learning. Existing research on privacy preservation can be categorized into two main approaches: encryption-based and perturbation-based methods.

Encryption-based methods encompass homomorphic encryption \citep{aono2017privacypreserving, zhang2020batchcrypt}, secret sharing \citep{shamir1979how}, and secure multiparty computation methods \citep{mohassel2017secureml}. These approaches focus on safeguarding data privacy during transmission and preventing unauthorized access to the original data by employing encryption and decryption in each data exchange process. However, the frequent encryption and decryption operations reduce the decentralized training efficiency greatly. Although homomorphic encryption allows computation on ciphertext and retrieval of the computed plaintext with a single decryption operation, it imposes stringent requirements on the calculation method and the time it occupies cannot be overlooked.

Perturbation-based methods, such as differential privacy \citep{geyer2017differentially, hao2019towards, mcmahan2017learning} and additive perturbation \citep{chamikara2021privacy, hu2020personalized, liu2020adaptive} are utilized in studies to prevent attackers from inferring data privacy. These methods involve adding noise directly to gradient values or training datasets. Although these methods are straightforward and require minimal additional training time, weak noise can be easily mitigated by noise reduction algorithms \citep{kargupta2003privacy}, while strong noise significantly reduces the training efficiency of the global model.

In summary, both types of privacy-preserving algorithms face a specific challenge when implemented in decentralized training frameworks: how to control the decline of training accuracy within an acceptable range while ensuring the efficiency of encryption. Further investigation is needed in future studies to determine the appropriate perspective to adopt in specialized training scenarios.

\subsection{Stage-Level Malicious Behaviors Detection}
As stated in Section \ref{Sec-PotentialThreats}, attackers engaging in poisoning attacks and privacy inference attacks demonstrate distinct motivations, capabilities, and malicious behaviors, thereby resulting in substantial divergences in the security algorithms applied to these scenarios. Prior studies have elucidated the practicality of defense mechanisms against targeted attacks, such as eliminating backdoors from trained models. However, this strategy proves inadequately effective against untargeted attacks. Nonetheless, it is evident that both poisoning attacks pursue a shared goal: tampering with activation values or gradient values. Consequently, conventional iteration-level defense methods, for instance, resilient aggregation techniques tackling Byzantine problems in FL, cannot be directly utilized in decentralized training frameworks. Therefore, a direct and efficient defense approach involves the implementation of a detection algorithm to identify malicious behaviors at the stage level.

Regrettably, this issue has not received adequate attention in the existing literature. To address this problem, we propose employing redundant computation to detect any malicious tampering between stages. In Section \ref{Sec-CaseStudy}, we present a comprehensive case study to illustrate the effectiveness of this detection methodology. Despite the additional GPU storage space requirements and the resulting decrease in training efficiency, our approach's robust defense capability convincingly validates its potential for future research.

\subsection{Fast Recovery from Failures}
In the event of a hardware failure or a detected poisoning attack, it is crucial for the pipeline to recover promptly. A straightforward method is to restart this training iteration every time encountering malicious behaviors. Despite its feasibility, this restart method wastes the results obtained in the current training iteration and leads to prolonged idle time, as subsequent computing resources remain underutilized for an extended period. Ensuring the continuous operation of the pipeline and quickly recovering the original data are essential considerations for a robust decentralized training framework. In Section \ref{Sec-CaseStudy}, we present alternative solutions to minimize computing resource consumption while achieving swift recovery from failures or attacks.

\section{A Case Study}
\label{Sec-CaseStudy}
We present a case study to examine the vulnerability of decentralized training and introduce our robust training framework. We substantiate our findings with experimental evidence that demonstrates the potential of this threat to disrupt model convergence. Furthermore, we demonstrate the effectiveness of our robust training framework in mitigating this risk. For convenience, we suggest a decentralized training framework consisting of $K$ stages. Additionally, $M_i$ represents the sub-layer of the $i$-th stage.


\subsection{Threat Model and Attack Methods}
\label{section-4.1}
\begin{figure}[t]
    \centering
    \includegraphics[width=\linewidth,page=2]{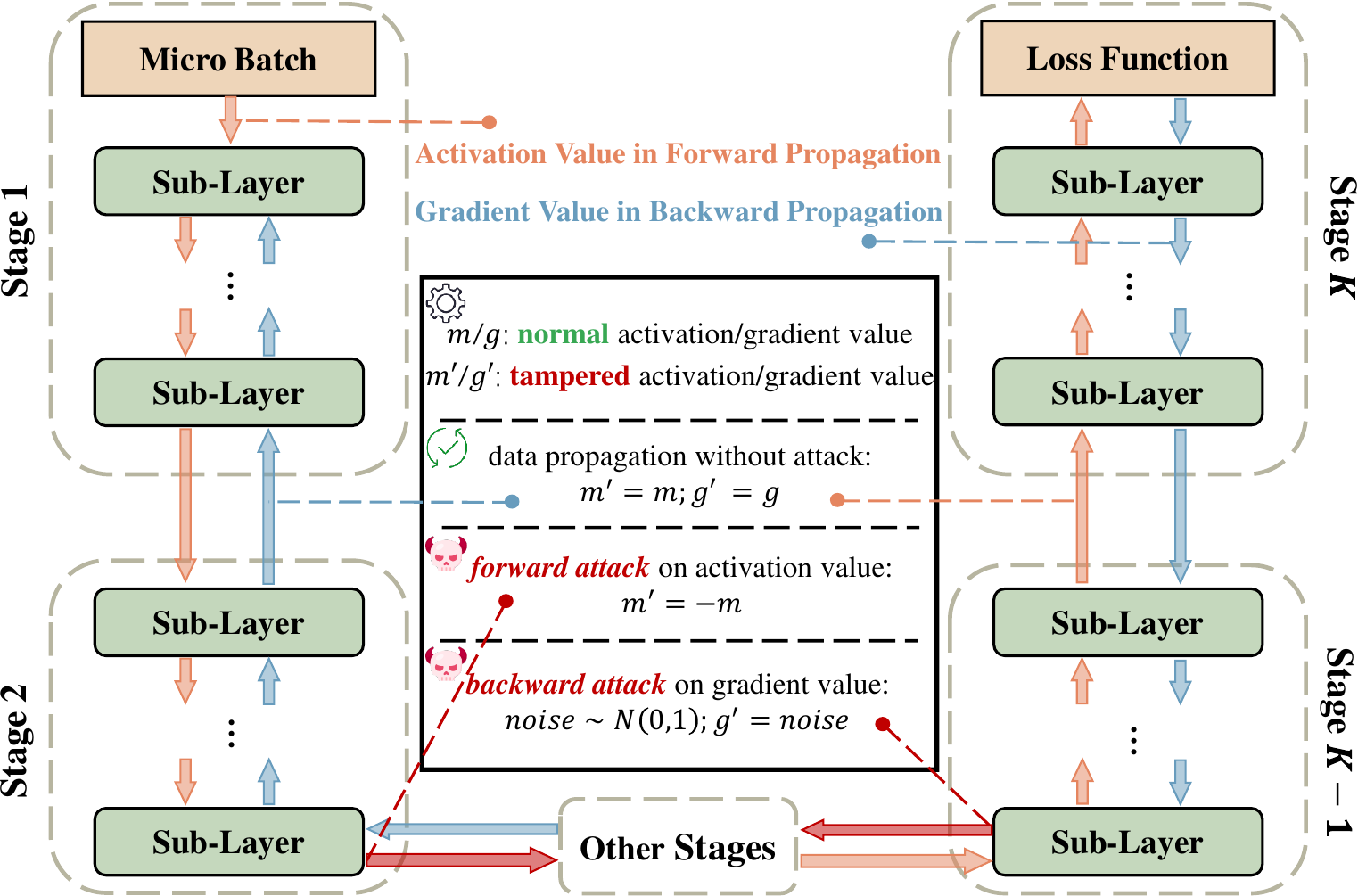}
    \caption{Illustration of the threat model in decentralized training with $K$ stages, depicting \textit{forward attack} and \textit{backward attack}. The orange arrows represent \textcolor{orange}{the transmission process of activation values during the forward propagation} while the blue arrows represent \textcolor{blue}{the transmission process of gradient values during the backward propagation.} The red arrows indicate \textcolor{red}{the location malicious behaviors occur}.}
    \label{figure-2}
    \vspace{-1 em}
\end{figure}

We assume an attacker, denoted as $\mathcal{A}$, who can randomly manipulate a stage, including both forward and backward propagation, during each iteration with a predetermined attack rate. If $\mathcal{A}$ successfully gains control of a stage, this particular stage transmits the malicious value $\mathbf{a}_{\text{out}}'$ to the subsequent stage, instead of the intended output $\mathbf{a}_{\text{out}}$, upon receiving a value $\mathbf{a}_{\text{in}}$. The malicious behavior during the model training process can be categorized as either a \textit{forward attack} or a \textit{backward attack}, as demonstrated in Figure \ref{figure-2} by orange and blue arrows, respectively. Notably, the initial and the final stages, which provide data and the corresponding labels, remain immune to attacks.

We employ two straightforward untargeted poisoning attack methods to simulate the actions of $\mathcal{A}$. In \textit{forward attack}, the malicious stage simply flips the sign of $\mathbf{a}_{\text{out}}$ resulting in $\mathbf{a}_{\text{out}}'=-\mathbf{a}_{\text{out}}$. In \textit{backward attack}, the malicious stage generates a Gaussian random variable $\mathbf{\phi} \sim N(0,1)$ with the shape as $\mathbf{a}_{\text{out}}$ and sets $\mathbf{a}_{\text{out}}'=\mathbf{\phi}$. Then the malicious stage sends $\mathbf{a}_{\text{out}}'$ to the next stage.


\subsection{Our Robust Training Framework}
\begin{figure*}[t]
    \centering
    \subfigure[detection strategy]{
    \label{figure-3-a}
    \includegraphics[width=0.5\textwidth,page=3]{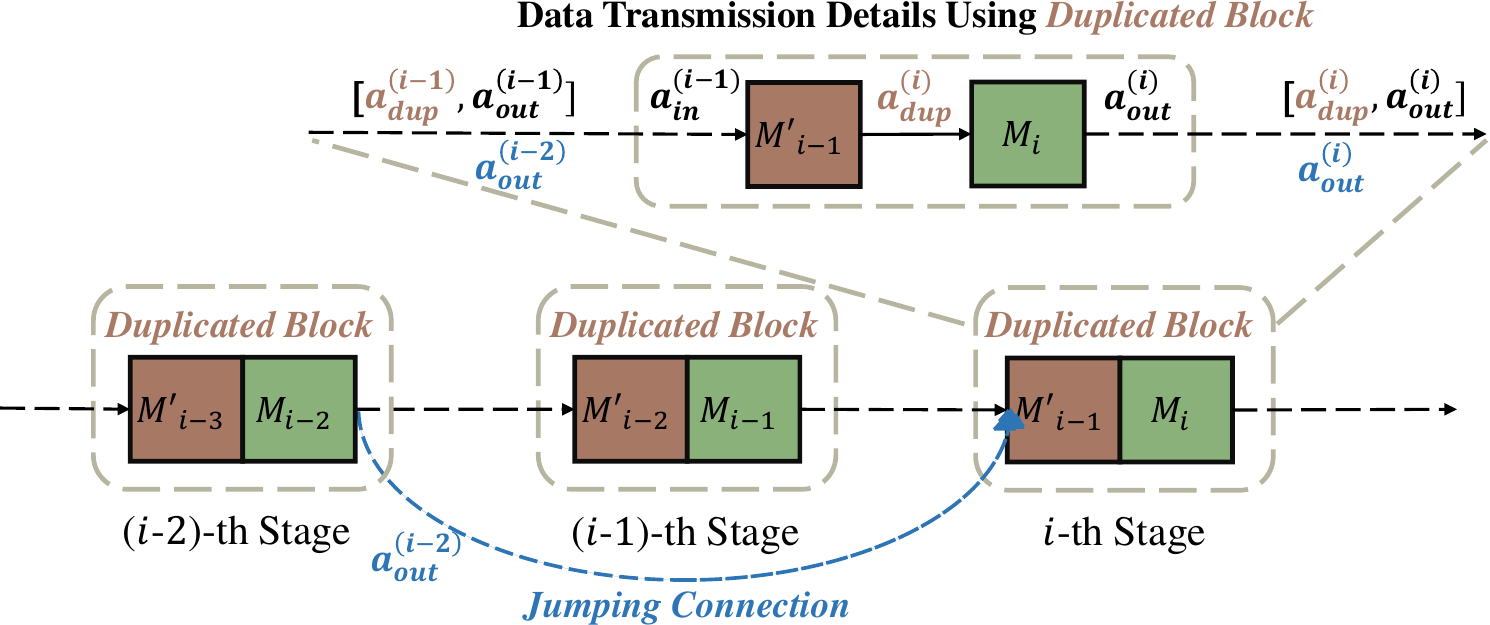}}
    \hspace{0.03\textwidth}
    \subfigure[efficient training framework]{
    \label{figure-3-b}
    \includegraphics[width=0.43\textwidth,page=4]{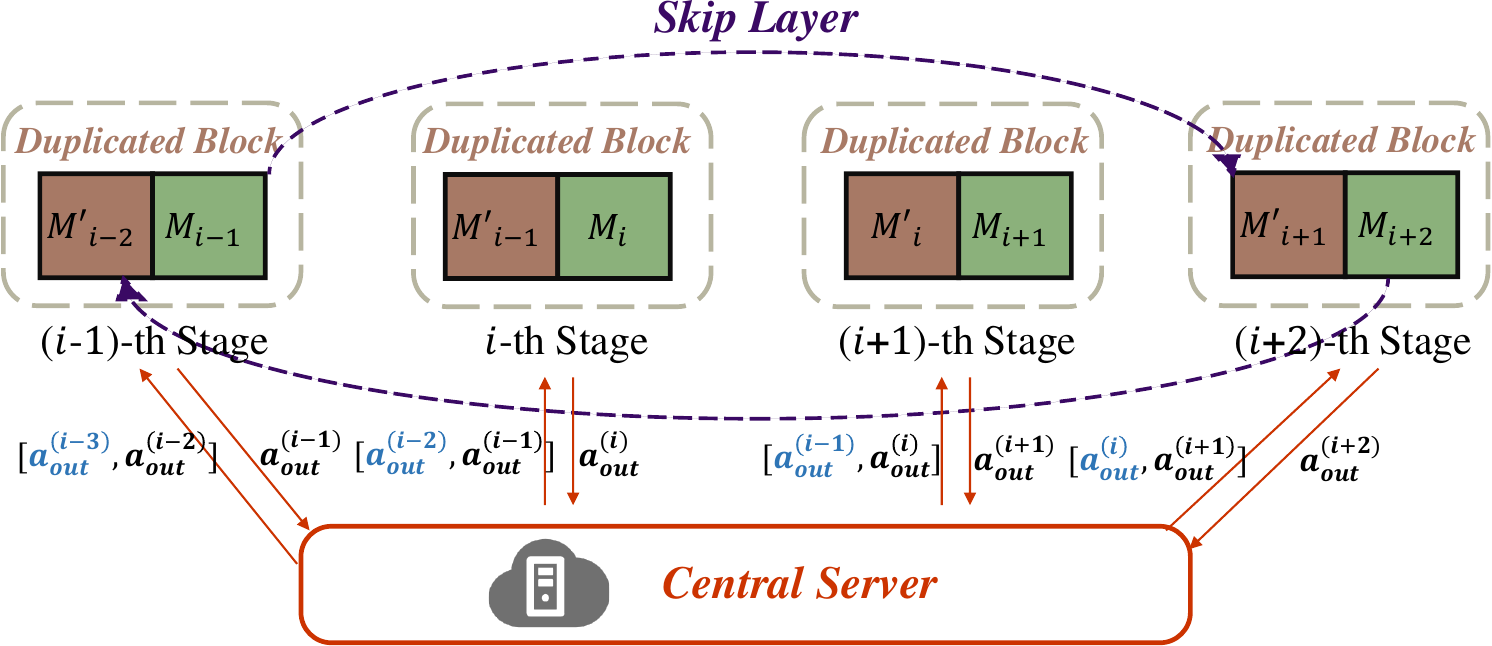}}
    \caption{Details about data transmission and structure of our proposed detection strategy and efficient training framework. For the detection strategy, the brown squares and green squares represent the \textcolor[RGB]{168,122,102}{duplicated layers} and the \textcolor[RGB]{139,177,123}{original layers}, respectively. And they together denote the duplicated block. Blue arrows represent the \textcolor[RGB]{46,117,182}{jumping connection} which is designed to detect a more knowledgeable attacker. For the efficient training framework, the red arrows represent the \textcolor[RGB]{204,51,0}{updated data transmission} while the purple arrow represents the \textcolor[RGB]{107,70,138}{data flow using skip layer}.}
    \label{figure-3}
\end{figure*}

Our robust training framework consists of two main components: attack detection and efficient training, depicted in Figure \ref{figure-3}. If our robust training framework does not detect malicious attacks, the training process continues as usual. However, in cases where malicious attacks are detected, the pipeline adopts the efficient training component to eliminate the bad consequences.

\noindent\textbf{Detection strategy.}
Naturally drawing the inspiration of redundant computation~\cite{patterson1988a, bogatyrev2015funtional} and Bamboo~\cite{thorpe2023bamboo}, we propose the \textit{duplicated block}. Taking the $i$-th stage as an example, it consists of redundant layers $M_{i-1}'$, duplicated from the $(i{-}1)$-th stage, and the original layers $M_i$ of the $i$-th stage.

During the forward propagation, the $i$-th stage sends its input $\mathbf{a}^{(i{-}1)}_{\text{out}}$ as $\mathbf{a}^{(i)}_{\text{dup}}$ and its output $\mathbf{a}^{(i)}_{\text{out}}$ to the next stage. Upon receiving data from the previous stage, the $i$-th stage first uses $M_{i-1}'$ to verify the compatibility between $\mathbf{a}^{(i-1)}_{\text{dup}}$ and $\mathbf{a}^{(i-1)}_{\text{out}}$. Only after this verification, the subsequent training process is performed. Once mismatched, the $i$-th stage triggers an alert and notifies the training initiator. Then the training initiator could take measures such as restarting this iteration and reusing the data sample.

To defend a more knowledgeable attacker and ensure the consistency of $\mathbf{a}^{(i-1)}_{\text{dup}}$ and $\mathbf{a}^{(i-2)}_{\text{out}}$, we introduce the \textit{jumping connection}. During each iteration, in addition to the aforementioned operations, the $i$-th stage transmits its output $\mathbf{a}^{(i)}_{\text{out}}$ to the $(i{+}2)$-th stage and receives $\mathbf{a}^{(i-2)}_{\text{out}}$ from the $(i{-}2)$-th stage. This verification invalidates a more stealthy attack that leverages an arbitrary input $\mathbf{a}_{\text{in}}'$ and sends the corresponding $\mathbf{a}_{\text{out}}'$ to the next stage. The verification and transmission process in the backward propagation mirrors the forward propagation but with the reversed data transmission direction. Throughout the entire training process, the parameters of $M_{i-1}$ and $M_{i-1}'$ remain identical. 

\noindent\textbf{Efficient training.}
If the $i$-th stage raises an alert, there will be a malicious stage among the $(i{-}2)$-th, $(i{-}1)$-th, and $i$-th stage. To narrow down the scope of suspicion, we introduce the \textit{central server}, which is immune to attacks, like the initial and final stages. Direct data transmission is no longer used across the stages. Instead, as demonstrated in Figure \ref{figure-3-b}, the $i$-th stage forwards output $\mathbf{a}^{(i)}_{\text{out}}$ to the central server. The central server subsequently sends the data pair $[\mathbf{a}^{(i-1)}_{\text{out}}, \mathbf{a}^{(i)}_{\text{out}}]$ to the $(i{+}1)$-th stage. All the verification and transmission behaviors inside the \textit{duplicated block} remain the same. Consequently, if the $(i{+}1)$-th stage raises an alert, the malicious stage is either the $i$-th or the $(i{+}1)$-th stage.

Inspired by stochastic depth \citep{huang2016deep, emin2018skip, hayou2021regularization}, we propose the \textit{skip layer} method to avoid restarting the training iteration when encountering attacks. Specifically, if the $(i{+}1)$-th stage raises an alert in the forward propagation, the central server will bypass the $i$-th and $(i{+}1)$-th stage, transmitting data directly between the $(i{-}1)$-th and $(i{+}2)$-th stage. To ensure consistency of parameters between the original and redundant layers, we keep $M_{i-1}$ and $M_{i+1}'$ unchanged while updating model parameters.



\subsection{Experiments}

\noindent\textbf{Experimental setup.}
We fine-tune multiple LLMs, including GPT-2 \citep{radford2019language}, Bloom \citep{workshop2022bloom}, and Opt \citep{zhang2022opt}, with different parameter sizes ranging from 345M to 7B. All the model checkpoints can be downloaded from HuggingFace.
We employ text-generation tasks on wikitext2, arxiv abstracts, and openwebtext datasets to conduct our evaluations. Our primary metric for assessing model performance is perplexity and GPipe \citep{huang2019gpipe} is used for our experiments as the base framework. To simulate the heterogeneous computing resources in real scenarios, the model is partitioned into six different computing resources, including A40, V100, RTX 3090, and Quadro RTX 5000. We utilize the clean model as the baseline to evaluate the vulnerability of the decentralized pipeline parallel training, and the attacked model to evaluate our robust training framework. We set the learning rate to $5\text{e}{-}6$ during training, and the batch size and micro-batch size to $4$ and $1$, respectively. In order to maintain consistency between the duplicated model and the main model, dropout is not employed in any of our experiments. 

\noindent\textbf{Vulnerability of decentralized pipeline parallel training.}
We first assess the vulnerability of different LLMs to \textit{forward attack} and \textit{backward attack}. Their influences on training accuracy are presented in Table \ref{tab1}. We denote the ratio of the attacked training iterations to the total number of training iterations as the attack rate. It is observed that the two attack methods yield excellent results when the attacking rate is set to 0.7. After sufficient training iterations without applying any defense measures, the perplexity of the model under backward attacks increases by at least sevenfold in comparison to that of the clean model, often extending to several tens of times. However, when the attack rate is set to 0.3, the attack's effectiveness is not consistently as good. We analyze that this discrepancy may arise from the lower probability of malicious behaviors, offset by the higher probability of normal values, which allows the model to converge with adequate training iterations. The absence of the dropout, leading to a severe overfitting phenomenon, could be a significant factor contributing to the subpar performance of the attacked model at an attack rate of 0.3.

\input{Tables/Table_1}

\noindent\textbf{Effectiveness of robust training framework.}
We demonstrate the effectiveness of our robust training framework when employing the \textit{forward attack} with the attack rate set to 0.5 in Table \ref{tab2}. We denote the perplexity on the clean model, the attacked model without any defense, and the attacked model under our robust training framework as clean, attack, and ours, respectively.

We observe that the perplexity of our model can improve up to 102.2 times compared to the perplexity of the attacked model when using the robust training framework. What's more, models using this framework even exhibit lower perplexity than the original models. Even when assessing the perplexity of Bloom-560M on arxiv, we note that the models employing our robust framework have only one-third perplexity of the original model. We speculate that this finding is consistent with the anomalous results presented in Table \ref{tab1}. Remarkably, in the absence of employing the dropout parameter, the skip layer acts as a highly effective regularization technique, mitigating the overfitting phenomenon of models.

\input{Tables/Table_2}

\section{Conclusion}
This paper primarily explores the robustness of decentralized training frameworks utilizing pipeline parallelism for training LLMs. Initially, we identify and classify the potential threats, including hardware failures, privacy inference attacks, and poisoning attacks, based on the attackers' objectives and capabilities. We then compare the structural differences between decentralized pipeline parallel training and FL. Additionally, we analyze the inherent reasons why existing security methods cannot be directly applied to the decentralized training frameworks. Following this, we propose a vision for a secure and robust framework for decentralized training. Lastly, we illustrate the vulnerability of the decentralized pipeline parallel training framework through a concrete case study and introduce an attack detection method, as well as an efficient training framework. Through experiments, we confirm that conventional decentralized training frameworks are vulnerable to attacks, and our approach effectively enhances its security. 
We anticipate that this paper can raise awareness of security concerns and contribute to enhancing the safety and reliability of decentralized training for LLMs.



\bibliographystyle{ACM-Reference-Format}
\bibliography{main}

\end{document}

%% file: Tables/Table_1.tex
\begin{table}[t]
\caption{Vulnerability of pipeline parallelism in decentralized training of LLMs on three datasets and two attack rates.}
\label{tab1}
\centering
\resizebox{0.98\linewidth}{!}{

\begin{tabular}{cc|c|cc|cc}
\hline
\multicolumn{2}{c|}{\textbf{attack methods \& attack rates→}} &
  \multicolumn{1}{l|}{\multirow{2}{*}{clean}} &
  \multicolumn{2}{c|}{\textit{forward attack}} &
  \multicolumn{2}{c}{\textit{backward attack}} \\ \cline{1-2} \cline{4-7} 
\multicolumn{2}{c|}{\textbf{LLM \& datasets↓}}                & \multicolumn{1}{l|}{} & 0.3    & 0.7     & 0.3    & 0.7     \\ \hline
\multicolumn{1}{c|}{\multirow{3}{*}{Opt-350M}}  & wikitext    & 29.77                 & 24.82  & 52.37   & 27.73  & 2128.31 \\ \cline{2-7} 
\multicolumn{1}{c|}{}                           & arxiv       & 22.61                 & 20.90  & 1383.81 & 56.14  & 1384.22 \\ \cline{2-7} 
\multicolumn{1}{c|}{}                           & openwebtext & 41.38                 & 38.30  & 3578.41 & 355.31 & 3584.42 \\ \hline
\multicolumn{1}{c|}{\multirow{3}{*}{GPT2-1.5B}} & wikitext    & 40.05                 & 56.43  & 2503.65 & 25.454 & 788.4   \\ \cline{2-7} 
\multicolumn{1}{c|}{}                           & arxiv       & 35.34                 & 28.89  & 843.38  & 23.42  & 275.4   \\ \cline{2-7} 
\multicolumn{1}{c|}{}                           & openwebtext & 53.41                 & 988.80 & 3226.01 & 104.87 & 2064.69 \\ \hline
\end{tabular}
}
\end{table}

%% file: Tables/Table_2.tex
\begin{table}[t]
\caption{Effectiveness of our robust training framework compared to clean and attacked model without any defense.}
\label{tab2}
\centering
\resizebox{0.98\linewidth}{!}{
\begin{tabular}{c|ccc|ccc}
\hline
\textbf{datasets \& modes→} & \multicolumn{3}{c|}{arxiv} & \multicolumn{3}{c}{openwebtext} \\ \hline 
\textbf{models↓}           & clean  & attack   & ours  & clean     & attack    & ours    \\ \hline
Opt-350M          & 22.61  & 601.92   & 19.31  & 41.38     & 3563.56   & 34.85   \\ \hline
Bloom-560M        & 67.15  & 1682.94  & 22.54  & 122.25    & 3984.84   & 61.65   \\ \hline
GPT2-1.5B         & 35.34  & 185.11   & 19.12  & 53.41     & 2435.17   & 36.56   \\ \hline
Bloom-7B          & 59.06  & 818.43   & 27.91  & 102.94    & 3077.24   & 52.62   \\ \hline
\end{tabular}
}
\end{table}